\pdfoutput=1
\documentclass[11pt]{article}
\usepackage{acl}

\usepackage{times}
\usepackage{latexsym}
\usepackage[T1]{fontenc}
\usepackage[utf8]{inputenc}
\usepackage{microtype}
\usepackage{inconsolata}

\usepackage{kotex}
\usepackage{multirow}
\usepackage{amsmath}
\usepackage{amsfonts}
\usepackage{graphicx}
\usepackage{cleveref}
\crefformat{section}{\S#2#1#3}
\crefformat{subsection}{\S#2#1#3}
\crefformat{subsubsection}{\S#2#1#3}
\usepackage{graphicx}
\usepackage{dblfloatfix} 
\usepackage{subcaption}
\usepackage{booktabs}
\usepackage[toc,page]{appendix}
\usepackage[export]{adjustbox}
\usepackage{color}
\usepackage{hyperref}

\author{Young Hyun Yoo$^{\dagger}$, Jii Cha$^{\dagger}$, Changhyeon Kim, Taeuk Kim$^*$ \\
Hanyang University, Seoul, Republic of Korea \\
{\tt \{somebodil,skchajie,livex,kimtaeuk\}@hanyang.ac.kr}}

\title{Hyper-CL: Conditioning Sentence Representations with Hypernetworks}

\newcommand{\astfootnote}[2]{
    \let\oldthefootnote=\thefootnote
    \setcounter{footnote}{#1}
    \renewcommand{\thefootnote}{\fnsymbol{footnote}}\footnotetext{#2}
    \let\thefootnote=\oldthefootnote
}

\begin{document}
\maketitle
\begin{abstract}
While the introduction of contrastive learning frameworks in sentence representation learning has significantly contributed to advancements in the field, it still remains unclear whether state-of-the-art sentence embeddings can capture the fine-grained semantics of sentences, particularly when conditioned on specific perspectives.
In this paper, we introduce Hyper-CL, an efficient methodology that integrates hypernetworks with contrastive learning to compute conditioned sentence representations.
In our proposed approach, the hypernetwork is responsible for transforming pre-computed condition embeddings into corresponding projection layers. 
This enables the same sentence embeddings to be projected differently according to various conditions.
Evaluation of two representative conditioning benchmarks, namely conditional semantic text similarity and knowledge graph completion, demonstrates that Hyper-CL is effective in flexibly conditioning sentence representations, showcasing its computational efficiency at the same time.
We also provide a comprehensive analysis of the inner workings of our approach, leading to a better interpretation of its mechanisms. 
Our code is available at \href{https://github.com/HYU-NLP/Hyper-CL}{https://github.com/HYU-NLP/Hyper-CL}.
\end{abstract}

\astfootnote{2}{Equal contribution. $^*$Corresponding author.}
\setcounter{footnote}{0}

\begin{figure}[t]
    \centering
    \includegraphics[width=\columnwidth]{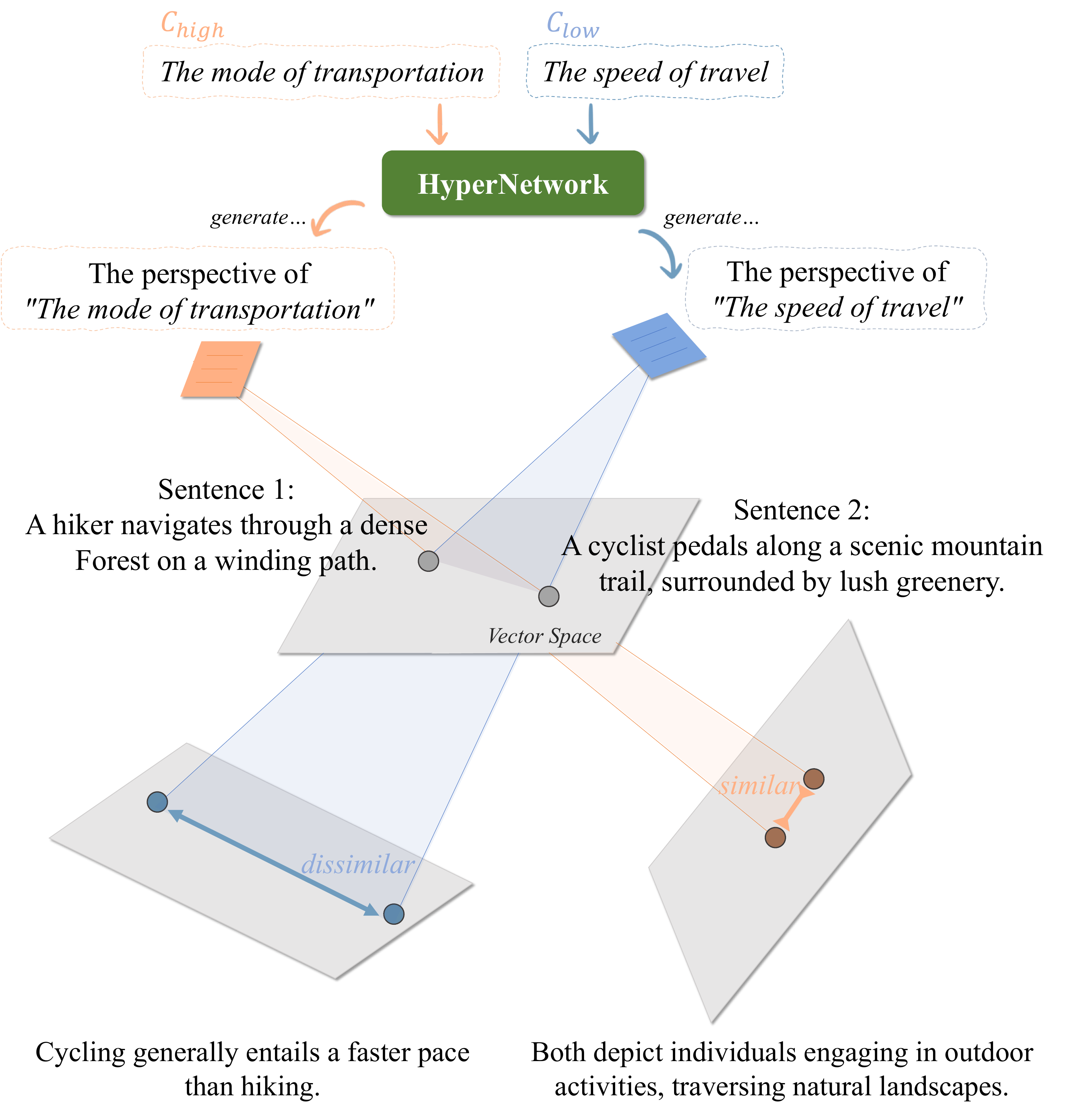}
    \caption{Illustration of our approach dubbed Hyper-CL. 
    In the example, two sentences are provided along with two distinct conditions, $c_{high}$ and $c_{low}$.
    Specifically, $c_{high}$ (\textcolor{orange}{orange}) denotes a condition that results in the sentences being interpreted more similarly, whereas $c_{low}$ (\textcolor{blue}{blue}) leads to a perspective in which the two sentences are understood as being relatively more distinct.
    The identical pair of sentences are projected into different subspaces that reflect the provided conditions.
    }
    \label{fig:fig_illustration_hyperCL}
\end{figure}

\section{Introduction} \label{introduction}

Building upon the established correlation between language model performance and computational capacity \cite{kaplan2020scaling}, there has emerged an undeniable trend towards the adoption of ever-larger language models across a diverse range of NLP applications.
This trend is also evident in the computation of sentence or text representations. 
Despite the ongoing popularity of compact encoders such as BERT \cite{devlin-etal-2019-bert} and RoBERTa \cite{liu2019roberta}, there is a growing inclination to leverage the capabilities of recent, larger language models, e.g., LLaMA-2 \cite{touvron2023llama}, even breaking from the conventional roles of encoders and decoders.
Consequently, the enduring challenge of finding a balance between performance and computational cost---a persistent issue in sentence representation learning \cite{reimers-gurevych-2019-sentence}---continues to be elusive.

\begin{figure*}[t!]
\centering
\begin{subfigure}{0.15\textwidth}
\includegraphics[width=\linewidth]{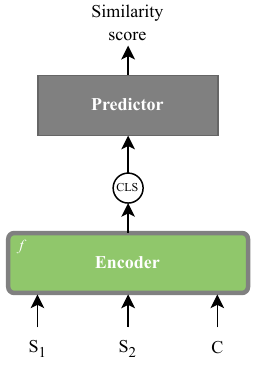}
\caption{Cross-Encoder}
\label{fig:fig_cross-enc}
\end{subfigure}\hspace{3mm}
\begin{subfigure}{0.185\textwidth}
\includegraphics[width=\linewidth]{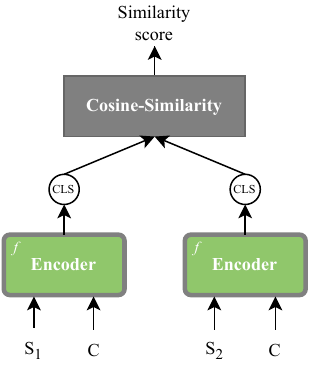}
\caption{Bi-Encoder}
\label{fig:fig_bi-enc}
\end{subfigure}\hspace{3mm}
\begin{subfigure}{0.26\textwidth}
\includegraphics[width=\linewidth]{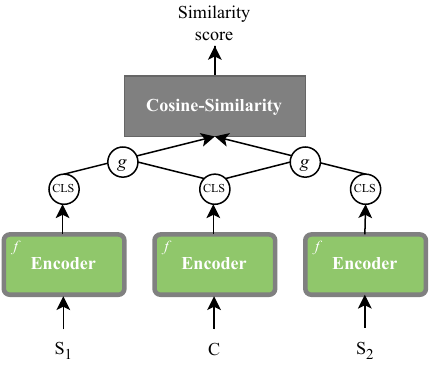}
\caption{Tri-Encoder}
\label{fig:fig_tri-enc}
\end{subfigure}\hspace{3mm}
\begin{subfigure}{0.26\textwidth}
\includegraphics[width=\linewidth]{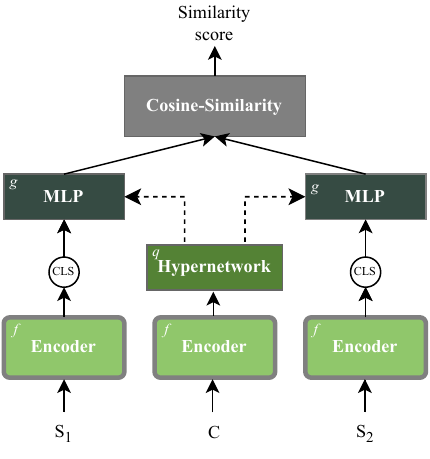}
\caption{Hyper-CL (Ours)}
\label{fig:fig_hyperCL}
\end{subfigure}
\caption{Four different types of architectures applicable for conditioning tasks. 
They utilize the [CLS] token embeddings from the encoder as representations of inputs.
From left to right: the \textit{cross-encoder} architecture encodes a triplet containing two sentences ($s_1, s_2$) and a condition ($c$) as a whole.
In the \textit{bi-encoder} setting, two sentence-condition pairs ($s_1, c$) and ($s_2, c$) are processed individually. 
The \textit{tri-encoder} configuration regard $s_1$, $s_2$, and $c$ as independent and encode them separately, followed by extra merging operations (e.g., Hadamard product). 
Finally, \textbf{Hyper-CL} resembles the \textit{tri-encoder}, but innovatively incorporates a hypernetwork responsible for constructing projection matrices to condition sentences $s_1$ and $s_2$, based on the embedding of the condition $c$.}
\label{fig:fig_architectures4}
\end{figure*}

In recent years, there has been a marked improvement in the quality of sentence embeddings, a progress primarily driven by the advent of contrastive learning frameworks (\citealp{kim-etal-2021-self,gao-etal-2021-simcse,chuang-etal-2022-diffcse}; \textit{inter alia}).
However, since the performance of these embeddings is generally evaluated based on their ability to encapsulate the overall meaning of the corresponding sentences---as measured by benchmarks like STS-B \cite{agirre-etal-2012-semeval,cer-etal-2017-semeval} and MTEB \cite{muennighoff-etal-2023-mteb}, it remains uncertain whether they adequately capture information relating to the various aspects of the source sentences.

For instance, consider the sentences (1) \textit{``A cyclist pedals along a scenic mountain trail, surrounded by lush greenery''} and (2) \textit{``A hiker navigates through a dense forest on a winding path, enveloped by the tranquility of nature''}. 
In terms of \textit{``The mode of transportation''}, these sentences should be perceived as similar since both depict individuals engaging in outdoor activities, traversing natural landscapes.
However, regarding \textit{``The speed of travel''}, they should be differentiated, as cycling generally entails a faster pace than hiking.
\citet{deshpande-etal-2023-c} reported that current models for sentence embeddings face challenges in recognizing the fine-grained semantics within sentences.
In other words, the existing models struggle to accurately detect the subtle shifts in sentence nuances that occur when conditioned on specific criteria.

In the literature, three prevalent approaches have been established for constructing \textit{conditioned} representations \cite{deshpande-etal-2023-c}, particularly in estimating their similarity (see Figure \ref{fig:fig_architectures4}).
The first is the \textit{cross-encoder} approach, which encodes the concatenation of a pair of sentences ($s_1$, $s_2$) with a condition ($c$), i.e., $[s_1; s_2; c]$.\footnote{$[\cdot; \cdot]$ represents the concatenation operation.}
The second method is the \textit{bi-encoder} architecture, computing separate representations of sentences $s_1$ and $s_2$ with the condition $c$---$[s_1; c]$ and $[s_2; c]$.
Despite their simplicity, both approaches share a notable limitation: the representation must be computed for every unique combination of sentences plus a condition.

On the other hand, the \textit{tri-encoder} architecture utilizes pre-computed embeddings of sentences $s_1$ and $s_2$ along with the condition $c$. 
It then employs a separate composition function responsible for merging the semantics of the sentence and condition.
Considering that the embeddings for each component can be cached and reused, this approach offers enhanced long-term efficiency.
The \textit{tri-encoder} architecture, despite its potential, falls short in performance compared to the \textit{bi-encoder}. 
This is primarily due to its inherent limitation, which is the inability to model explicit interactions between sentences and conditions during the representation construction process.
Therefore, there is a need to propose a revised version of the \textit{tri-encoder} architecture that improves its functionality without substantially sacrificing its efficiency.

In this work, we present \textbf{Hyper-CL}, a method that integrates \textbf{Hyper}networks \cite{ha2017hypernetworks} with \textbf{C}ontrastive \textbf{L}earning to efficiently compute conditioned sentence representations and their similarity. 
As illustrated in Figure \ref{fig:fig_hyperCL}, our proposed approach is derived from the \textit{tri-encoder} architecture. 
It introduces an additional hypernetwork tasked with constructing a condition-sensitive network on the fly. 
This network projects the original sentence embeddings into a specific condition subspace.
Figure \ref{fig:fig_illustration_hyperCL} illustrates the effectiveness of Hyper-CL in dynamically conditioning pre-computed sentence representations according to different perspectives.

We demonstrate the effectiveness of Hyper-CL by significantly reducing the performance gap with the \textit{bi-encoder} architecture in the Conditional Semantic Textual Similarity (C-STS) and Knowledge Graph Completion (KGC) tasks.
In particular, for C-STS, Hyper-CL demonstrates an improvement of up to 7.25 points in Spearman correlation compared to the original \textit{tri-encoder} architecture.
Furthermore, compared to the \textit{bi-encoder} approach, our method shows superior efficiency by reducing the running time by approximately 40\% on the C-STS dataset and 57\% on the WN18RR dataset.

\section{Background and Related Work} \label{sec:Background and Related Work}

In this paper, \textit{conditioning} refers to the presence of two or more signals, each represented as a natural language expression $c$.
These signals impact the interpretation of a sentence $s$, highlighting a specific aspect of the sentence \cite{galanti2020modularity}.
Here, we describe two representative tasks that involve conditioning, along with an introduction to the concept of hypernetworks.

\textbf{Conditional Semantic Textual Similarity (C-STS)} \cite{deshpande-etal-2023-c} is a task composed of four elements in one quadruplet: two sentences $s_1$ and $s_2$, a condition $c$ to consider when calculating the similarity between the two sentences, and a similarity score $y$.
Unlike the original Semantic Textual Similarity (STS) dataset \cite{agirre-etal-2012-semeval,cer-etal-2017-semeval}, C-STS computes similarity scores for the same sentence pair $s_1$ and $s_2$ under two distinct conditions $c_{high}$ and $c_{low}$.
The similarity scores of the conditioned sentences are expected to be high for $c_{high}$ and low for $c_{low}$.
Therefore, a model for this task is required to compute distinct representations for the same sentence under two different viewpoints.
To this end, a few basic architectures illustrated in Figure \ref{fig:fig_architectures4} have been proposed by \citet{deshpande-etal-2023-c}.
Our objective is to present a revision to the previous methods, pursuing the balance between performance and computational efficiency.

\textbf{Knowledge Graph Completion (KGC)} is the task focused on automatically inferring missing relationships or entities in a knowledge graph.
The knowledge graph is represented as a set of triplets ($h, r, t$), consisting of a head entity $h$, a relation $r$, and a tail entity $t$.
Link prediction, a subtask in KGC,\footnote{While this work focuses solely on link prediction, note that other relevant tasks exist within the domain of KGC.} aims to uncover unestablished yet plausible and novel relationships between entities \cite{bordes2013translating,toutanova2015observed}.
When given a head entity and a relation, the task of identifying the most suitable tail entity is known as head entity prediction. Conversely, the task of determining the appropriate head entity when a tail entity and relation are provided is termed tail entity prediction.

While two types of methodologies are generally available for KGC---embedding-based methods and text-based methods---our primary focus is on text-based methods that rely on the processing of textual information by language models.
We further categorize them into three types: \textit{cross-encoder}, \textit{encoder-decoder}, and \textit{bi-encoder} architectures.
Approaches such as KG-BERT \cite{yao2019kg} and MTL-KGC \cite{kim-etal-2020-multi}, which concatenate all triple elements (i.e., $[h; r; t]$), are classified as \textit{cross-encoder}.
Methods like StAR \cite{wang2021structure} and SimKGC \cite{wang-etal-2022-simkgc}, which separately embed $[h; r]$ and $t$ in tail prediction tasks, are classified as \textit{bi-encoder}. 
Lastly, GenKGC \cite{xie2022discrimination} and KG-S2S \cite{chen-etal-2022-knowledge}, which directly generate tail entity text based on the remaining $[h; r]$, are classified as \textit{encoder-decoder}.

On the other hand, \textbf{hypernetworks} refer to a type of neural network that generates the weights or parameters for another neural network, known as the primary network \cite{ha2017hypernetworks,chauhan2023brief,majumdar2023hyperlora}.
In essence, a hypernetwork enables the dynamic construction of the primary network, allowing its function to adapt flexibly based on the input or condition.
For instance, \citet{galanti2020modularity} demonstrate that, even with a compact primary network, hypernetworks can effectively learn and apply diverse functions for various inputs, provided the hypernetwork itself is sufficiently large.
In our settings, we also endeavor to harness the advantages of hypernetworks, ensuring that conditioned sentence representations are dynamically computed and adapted in response to changing conditions.

\begin{figure}[t]
    \centering
    \begin{adjustbox}{minipage=\columnwidth,scale=0.95}
    \includegraphics[width=\linewidth, height=7cm]{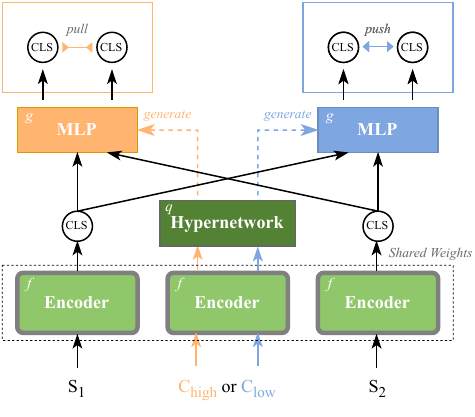}
    \end{adjustbox}
    \caption{Training procedure of Hyper-CL.
    It introduces a hypernetwork $q$ to construct the weights of multi-layer perceptrons (MLPs), i.e., $g$, based on the condition. 
    The MLPs are then used to project sentence embeddings onto subspaces, resulting in condition-aware sentence embeddings. 
    Hyper-CL is trained with a contrastive objective, utilizing pairs of condition-aware sentence embeddings, one with a high condition $c_{high}$ and the other with a low condition $c_{low}$.
    Note that every embedding is the output of the same encoder $f$.}
    \label{fig:fig_framework}
\end{figure}

\section{Proposed Method: Hyper-CL} \label{proposed method: hyper-cl}

\subsection{Motivation} 
As briefly mentioned in \cref{introduction}, current approaches to sentence conditioning demonstrate a clear trade-off between performance and computational efficiency.
Approaches that enable direct interaction between a sentence and a condition within encoders—the \textit{cross-encoder} and \textit{bi-encoder} architectures—gain enhanced performance but at the cost of reduced efficiency.
In contrast, the \textit{tri-encoder} architecture, while enabling efficient conditioning with pre-computed sentence and condition embeddings, tends to have inferior performance compared to its counterparts.
Note that this trade-off becomes more pronounced as the number of sentences and conditions to be processed increases.

Formally, consider a computationally intensive language model-based encoder $f$, and a less demanding composition network $g$, which is required in the case of the \textit{tri-encoder} architecture.
In terms of the \textit{bi-encoder}, obtaining the set of every possible conditioned embedding $\mathcal{H} = \{ h_{sc} \vert \forall s \forall c, s \in \mathcal{S}, c \in \mathcal{C} \}$ necessitates $|\mathcal{S}| \times |\mathcal{C}|$ repetitions of the heavy operation imposed by $f$, where $|\mathcal{S}|$ denotes the total number of sentences and $|\mathcal{C}|$ represents the total number of possible conditions.
In contrast, the \textit{tri-encoder} architecture only requires heavy operations by $f$ for each sentence \(s\) and condition \(c\) just once.
This implies that only \(|\mathcal{S}| + |\mathcal{C}|\) heavy operations are needed, followed by \(|\mathcal{S}| \times |\mathcal{C}|\) lightweight operations by $g$ to obtain the conditioned embeddings.
As a result, the \textit{tri-encoder} and its variants become more efficient if the cost of computing $g$ is markedly lower than that of $f$, thereby amortizing the cost for computing $\mathcal{H}$.

In this work, we aim to develop a new architecture for sentence conditioning that inherits the efficiency merits of the \textit{tri-encoder} architecture, while simultaneously outperforming the original in terms of performance. 
To achieve this, we propose the use of hypernetworks to implement $g$, facilitating the dynamic construction of conditioning networks while maintaining reasonable cost-efficiency.

\subsection{Framework and Training Procedure} \label{framework and training procedure}

The framework of Hyper-CL and its training procedure are listed as follows (also see Figure \ref{fig:fig_framework}):
\begin{enumerate}
\item First, it computes the embeddings of a sentence $s$ and a condition $c$ using the same embedding model $f$: $h_{s}=f(s)$ and $h_{c}=f(c)$.
\item Given the condition embedding $h_c$, a hypernetwork $q:\mathbb{R}^{N_{h}}\rightarrow\mathbb{R}^{N_{h}\times{N_{h}}}$ outputs a linear transformation matrix $W_c$ for conditioning: $W_c=q(h_c)$.\footnote{$N_{h}$: The dimensionality of sentence embeddings, which is equal to the size of the hidden states of an encoder model.}
\item We encode the condition-aware sentence embedding $h_{sc}$ based on the matrix $W_c$ and the sentence embedding $h_s$: $h_{sc}=W_c\cdot{h_s}.$
\item For training, we perform contrastive learning with the conditional sentence embeddings $h_{sc}$, whose details are explained in the following.
\end{enumerate}

\subsection{Contrastive Learning in Subspaces} \label{contrastive Learning in condition subspaces}

The conditioning network composed of $W_c$ is a linear neural network.
In other words, it can be interpreted as a linear transformation function $g:\mathbb{R}^{N_h}\rightarrow{\mathbb{R}^{N_h}}$, mapping from the original semantic space of sentence embeddings to a specialized condition subspace.
We demonstrate that conducting contrastive learning within the subspace of specific viewpoints yields greater effectiveness compared to performing the same process in the general space.
We apply separate task-oriented contrastive learning objectives for the tasks, C-STS and KGC.

\paragraph{C-STS} The C-STS task entails providing conditions $c_{high}$ and $c_{low}$ for two sentences, $s_1$ and $s_2$. 
This setup induces different interpretations of the relationship between the two sentences---one being more similar under $c_{high}$ and the other more dissimilar under $c_{low}$.
In a given instance from the dataset, Hyper-CL generates two pairs of conditioned sentence embeddings, i.e., $(h_{s_{1}c_{high}},h_{s_{2}c_{high}})$ and $(h_{s_{1}c_{low}},h_{s_{2}c_{low}})$.
Since these pairs correspond to positive and negative pairs in the contrastive objective, we directly utilize them for training.

Considering that the C-STS dataset already contains gold-standard similarity values for the two sentences under each condition, it seems reasonable to employ the Mean Squared Error (MSE) objective in conjunction with contrastive learning. 
However, as training progresses, we can speculate that MSE objectives that utilize labels will provide relatively more fine-grained granularity compared to contrastive objectives that do not.
Therefore, to mitigate the relatively strong impact of the contrastive objective, we apply the InfoNCE \cite{oord2018representation} loss with high temperature, as follows:

\begin{equation*} \label{eq:eq1}
    \resizebox{.99\linewidth}{!}{
    $L_{CL} = -\log \frac{e^{{\phi(h_{s_{1}c_{high}}, h_{s_{2}c_{high}})/\tau}}}{e^{{\phi(h_{s_{1}c_{high}}, h_{s_{2}c_{high}})/\tau}} + e^{{\phi(h_{s_{1}c_{low}}, h_{s_{2}c_{low}})/\tau}}},$}
\end{equation*}
where $\phi$ is the cosine similarity function and $\tau$ is a temperature hyperparameter.
The MSE objective is as follows:
\begin{equation*} \label{eq:eq2}
L_{MSE} = \lVert \phi(h_{s_{1}c}, h_{s_{2}c}) - y \rVert_2^2,
\end{equation*}
where $c$ can be either $c_{high}$ or $c_{low}$.
By combining the two above formulas, the final form of our training objective for C-STS becomes:
\begin{equation*}
L = L_{MSE} + L_{CL}.
\end{equation*}
Note that $L$ is averaged over data instances in the training set.

\paragraph{KGC} We follow the setting of SimKGC \cite{wang-etal-2022-simkgc}, except that we leverage Hyper-CL instead of the \textit{bi-encoder} architecture.
For each triplet of head entity, relation, and tail entity ($h, r, t$), we treat entities as sentences and relations as conditions, framing KGC as a conditioning task.
Furthermore, given that Hyper-CL is simple and flexible enough to adopt various techniques from SimKGC, such as the use of self-negative, pre-batch negatives, and in-batch negatives, we decide to apply these tricks to our method as well.
In conclusion, the final training objective for KGC is as follows (refer to \citet{wang-etal-2022-simkgc} for more details):
\begin{equation*}
\resizebox{.99\linewidth}{!}{
    $L_{CL} = -\log \frac{e^{(\phi(h_{hr}, h_{t}) - \gamma)/\tau}}{e^{(\phi(h_{hr}, h_{t}) - \gamma)/\tau} + \sum_{j=1}^{N} e^{\phi(h_{hr}, h_{{t'}})/\tau}}$}
\end{equation*}
where $h_{hr}$, $h_{t}$ and $h'_{t}$ are the relation-aware head embedding, tail embedding, random embedding (i.e., self-negative, pre-batch negative, and in-batch negative), respectively. 
The relation-aware head embedding corresponds to a conditional sentence embedding.
$\gamma$ is an additive margin, \(\phi\) is the cosine similarity, and $\tau$ is a learnable parameter.

\begin{table*}[t!]
\begin{center}\small
\begin{tabular}{lccc}
\toprule[1.5pt]
\textbf{Method} & \textbf{\# Params} & \textbf{Spearman} & \textbf{Pearson}\\
\midrule
\multicolumn{4}{l}{\textit{tri-encoder architectures}} \\
\midrule
DiffCSE\(^{\dag}_{base}\) & 125M & 28.9\(_{0.8}\) & 27.8\(_{1.2}\) \\
\(*\)DiffCSE\(_{base+hyper64\text{-}cl}\) & 200M & 33.10\(_{0.2}\) & 31.68\(_{0.6}\) \\
\(*\)DiffCSE\(_{base+hyper\text{-}cl}\) & 578M & \textbf{33.82}\(_{0.1}\) & \textbf{33.10}\(_{0.3}\)\\
\addlinespace
SimCSE\(^{\dag}_{base}\) & 125M & 31.5\(_{0.5}\) & 31.0\(_{0.5}\) \\
\(*\)SimCSE\(_{base+hyper64\text{-}cl}\) & 200M & 38.36\(_{0.1}\) & 37.53\(_{0.04}\) \\
\(*\)SimCSE\(_{base+hyper\text{-}cl}\) & 578M & \textbf{38.75}\(_{0.3}\) & \textbf{38.38}\(_{0.3}\) \\
\addlinespace
SimCSE\(^{\dag}_{large}\) & 355M & 35.3\(_{1.0}\) & 35.6\(_{0.9}\) \\
\(*\)SimCSE\(_{large+hyper85\text{-}cl}\) & 534M & 38.12\(_{1.4}\) & 37.47\(_{1.4}\) \\
\(*\)SimCSE\(_{large+hyper\text{-}cl}\) & 1431M & \textbf{39.60\(_{0.2}\)} & \textbf{39.96\(_{0.3}\)} \\
\midrule
\multicolumn{4}{l}{\textit{bi-encoder architectures}} \\
\midrule
DiffCSE$^{\dag}_{base}$ & 125M & 43.4$_{0.2}$ & 43.5$_{0.2}$\\
SimCSE$^{\dag}_{base}$ & 125M & 44.8$_{0.3}$ & 44.9$_{0.3}$\\
SimCSE$^{\dag}_{large}$ & 355M & \textbf{47.5$_{0.1}$} & \textbf{47.6$_{0.1}$}\\
\bottomrule[1.5pt]
\end{tabular}
\end{center}
\caption{Performance on C-STS measured by Spearman and Pearson correlation coefficients. 
The best results are in \textbf{bold} for each section. 
*: indicates the results of Hyper-CL. 
$\dag$: denotes results from \citet{deshpande-etal-2023-c}.
}
\label{tab:tab1}
\end{table*}

\begin{table*}[t]
\begin{center}\small
\begin{tabular}{l c c c c c c c c}
\toprule[1.5pt]
\multirow{2}{*}{\shortstack[c]{\textbf{Method}}} & \multicolumn{4}{c}{\textbf{WN18RR}} & \multicolumn{4}{c}{\textbf{FB15K-237}} \\
\cmidrule(lr){2-5}\cmidrule(lr){6-9}
 & MRR & Hits@1 & Hits@3 & Hits@10 & MRR & Hits@1 & Hits@3 & Hits@10 \\
\midrule
\multicolumn{9}{l}{\textit{cross-encoder architectures}} \\
\midrule
KG-BERT$^\dag$ & 0.216 & 0.041 & 0.302 & 0.524 & - & - & - & 0.420 \\
MTL-KGC$^\dag$ & 0.331 & 0.203 & 0.383 & 0.597 & 0.267 & 0.172 & 0.298 & 0.458 \\
\midrule
\multicolumn{9}{l}{\textit{encoder-decoder architectures}} \\
\midrule
GenKGC$^\dag$ & - & 0.287 & 0.403 & 0.535 & - & 0.192 & 0.355 & 0.439 \\
KG-S2S$^\dag$ & 0.574 & \underline{0.531} & 0.595 & 0.661 & \textbf{0.336} & \textbf{0.257} & \textbf{0.373} & \underline{0.498} \\
\midrule
\multicolumn{9}{l}{\textit{bi-encoder architectures}} \\
\midrule
StAR$^\dag$ & 0.401 & 0.243 & 0.491 & 0.709 & 0.296 & 0.205 & 0.322 & 0.482 \\
SimKGC\(^\ddag\) & \textbf{0.666} & \textbf{0.587} & \textbf{0.717} & \underline{0.800} & \textbf{0.336} & \underline{0.249} & \underline{0.362} & \textbf{0.511} \\
\midrule
\multicolumn{9}{l}{\textit{tri-encoder architectures}} \\
\midrule
SimKGC\(_{hadamard\text{-}product}\) & 0.164 & 0.004 & 0.243 & 0.481 & 0.153 & 0.092 & 0.162 & 0.274 \\
SimKGC\(_{concatenation}\) & 0.335 & 0.226 & 0.382 & 0.550 & 0.271 & 0.193 & 0.292 & 0.430 \\
*SimKGC\(_{hyper\text{-}cl}\) & \underline{0.616} & 0.506 & \underline{0.690} & \textbf{0.810} & \underline{0.318} & 0.231 & 0.344 & 0.496 \\
*SimKGC\(_{hyper64\text{-}cl}\) & 0.548 & 0.427 & 0.626 & 0.770 & 0.305 & 0.219 & 0.331 & 0.479 \\
\bottomrule[1.5pt]
\end{tabular}
\end{center}
\caption{Results on the WN18RR and FB15K-237 datasets for KGC, measured by MRR and Hits@K.
The best results are highlighted in \textbf{bold}, while the next best results are \underline{underlined} for each column.
*: indicates the results of applying Hyper-CL.
$\dag$: denotes results from \citet{chen-etal-2022-knowledge}. 
$\ddag$: denotes results from \citet{wang-etal-2022-simkgc}.
Other results are implemented and evaluated by the authors.}
\label{tab:tab2}
\end{table*}

\subsection{Optimization of Hypernetworks}

In the original formulation presented in \cref{framework and training procedure}, the number of parameters for the hypernetwork $ q:{\mathbb{R}^{N_h}}\rightarrow{\mathbb{R}^{N_h \times N_h}}$ is the cube of ${N_h}$, which could lead to cost inefficiency.
To address this issue, we propose decomposing the network into two low-rank matrices, drawing inspiration from low-rank approximation \cite{yu2017compressing,hu2021lora}.
In particular, we introduce two smaller hypernetworks of the same size: \(q_1:{\mathbb{R}^{N_h}}\rightarrow{\mathbb{R}^{N_h \times N_K}}\) and \(q_2:{\mathbb{R}^{N_h}}\rightarrow{\mathbb{R}^{N_h \times N_K}}\) to generate two low-rank matrices $W_{c_1} = q_{1}(h_c)$ and $W_{c_2} = q_{2}(h_c)$, where $N_k$ is much smaller than $N_h$.
We then obtain the final matrix through their product: \(W_c = W_{c1}W_{c2}^T\).\footnote{A similar method was proposed in \citet{majumdar2023hyperlora}.}

\subsection{Caching Conditioning Networks}

In the \textit{tri-encoder} architecture, once computed, the sentence and condition embeddings $h_s$ and $h_c$ can be cached and subsequently reused whenever conditioned embeddings related to them need to be computed.
Hyper-CL, building upon the \textit{tri-encoder} framework, not only inherits this advantage but also further improves time efficiency by caching the parameters of the entire conditioning networks $W_c = q(h_c)$ generated by the hypernetwork.
It is important to note that this approach is viable because the computation of the matrix $W_c$ depends solely on $h_c$, without requiring any other inputs.

\section{Experiments}

We apply Hyper-CL to various embedding models (i.e., encoders) and fine-tune them on the target task, denoting the resulting models with the subscript \textit{hyper-cl}.
If the rank of the hypernetwork ($N_k$) is different from $N_{h}$, we denote this value as \textit{hyperK-cl}.
We set $K$ as 64 and 85, for the base and large models respectively. 
A detailed explanation for the selection of $K$ can be found in the \cref{appendix:selection of k}.
We show the effectiveness of Hyper-CL by evaluating it on two downstream tasks.

\subsection{Conditional Semantic Textual Similarity}
We use DiffCSE \cite{chuang-etal-2022-diffcse} and SimCSE \cite{gao-etal-2021-simcse}, adaptations of RoBERTa \cite{liu2019roberta}, for the embedding model $f$.
Note that the key difference between the original \textit{tri-encoder} architecture and Hyper-CL lies in the composition network, $g$. 
The original uses the simple Hadamard product, while Hyper-CL employs hypernetworks to learn linear layers for this composition.

Table \ref{tab:tab1} summarizes the results of Hyper-CL in addition to baselines on C-STS.
Compared to the \textit{tri-encoder} baselines, Hyper-CL demonstrates improvements with up to a 7.25-point increase in Pearson correlation when based on SimCSE$_{base}$. 
This reduces the performance gap between the \textit{bi-encoder} and \textit{tri-encoder} from 13.3 to 6.05 points.

In addition, even when Hyper-CL is developed with low-rank approximation (i.e., $hyper64\text{-}cl$, $hyper85\text{-}cl$), its performance remains consistent. 
This indicates that the memory usage of hypernetworks can be effectively controlled, while both performance and time efficiency are maintained.

\subsection{Knowledge Graph Completion}
In KGC, the link prediction task entails computing relation-aware embeddings for head or tail entities and subsequently retrieving the top-K embeddings based on their similarity scores.
We consider two datasets for KGC: WN18RR \cite{bordes2013translating} and FB15k-237 \cite{toutanova2015observed}.

We employ text-based KGC models as baselines for evaluation.
Specifically, we use the SimKGC model that leverages all negatives (i.e., in-batch negative, pre-batch negative, and self-negative), which is also true when applying Hyper-CL to SimKGC.
We consider an extra baseline of applying the \textit{tri-encoder} architecture to SimKGC with different $g$---(1) Hadamard: performs the Hadamard product between the representations of a condition \(c\) and a sentence \(s\): \(g_1(h_c, h_s) = h_c \odot h_s\). (2) Concatenation: merge the two vectors, apply a dropout function, and halve the dimension using a linear layer: $g_2(h_c, h_s) = W \cdot d([h_c;h_s])$.

Table \ref{tab:tab2} presents the outcomes of our method and baselines on KGC, measured by MRR (Mean Reciprocal Rank) and Hits@K.
BERT$_{base}$ is leveraged for the embedding model $f$.
As a result, SimKGC$_{hyper\text{-}cl}$, representing the application of Hyper-CL to SimKGC, shows that there is no significant difference in performance compared to the original SimKGC, especially in terms of Hits@10.
Even for SimKGC$_{hyper64\text{-}cl}$, while there is a slight decrease in performance, it still yields competitive results and does not significantly trail behind other baselines.
Moreover, the performance of other methods in the \textit{tri-encoder} architecture falls significantly short of Hyper-CL's.
It is worth noting that our implementation is based on the \textit{tri-encoder} architecture, which guarantees significantly more efficiency in running time compared to the original SimKGC. The details of this analysis are in \cref{comparison of efficiency between bi-encoder and tri-encoder}.

\section{Analysis}

\subsection{Efficiency Comparison between Bi-Encoder and Tri-Encoder} \label{comparison of efficiency between bi-encoder and tri-encoder}

To assess the running time efficiency of Hyper-CL, enabled by its caching capability, we compare the execution time of our method with that of the \textit{bi-encoder} and \textit{tri-encoder} architectures.
We measure execution time, cache hit rate, and memory usage in scenarios with caching enabled.
Initially, the cache is empty, and embeddings are added to the cache upon each cache miss. 

Specifically, we estimate the inference time required for the C-STS and WN18RR datasets.
To simulate a realistic scenario where the number of data samples to be processed is substantial, we utilize all dataset splits (training, validation, and test) from the datasets.
For the C-STS dataset, we also evaluate the results of SimCSE\(_{hyper64\text{-}cl}\) as a solution for cases with large cache sizes.
For Hyper-CL, the transformation matrix $W_c$ is saved instead of embeddings. 
We set the batch size as 1 to minimize the overhead of cache storage and retrieval.

As observed in Table \ref{tab:tab3} and \ref{tab:tab4}, Hyper-CL reduces the running time by approximately 40\% on the C-STS dataset and 57\% on the WN18RR dataset than the \textit{bi-encoder} architecture. 
Compared to the na\"ive \textit{tri-encoder} architecture, Hyper-CL requires slightly more time, but we believe this is acceptable given its significantly improved performance.

In terms of cache hit rate, the \textit{tri-encoder} architecture, including Hyper-CL, achieves a much higher rate as it embeds each input separately, whereas the \textit{bi-encoder} architecture has a lower rate because both the sentence and condition combination must match to result in a cache hit.
Considering the significant gap in cache hit ratio between the \textit{bi-encoder} and \textit{tri-encoder} architectures, we expect that the efficiency of Hyper-CL will be more pronounced when deployed to process real-time streaming data from a large pool of users. 
In such scenarios, the diversity of input sentences and their conditions (relations) would be much higher than in our experimental settings, implying that the efficiency gap between the \textit{bi-encoder} and \textit{tri-encoder} architectures will be more severe.\footnote{The ratio of the number of unique items required to be cached by the \textit{tri-} and \textit{bi-encoder} architectures, i.e., \((|\mathcal{C}| + |\mathcal{S}|)/(|\mathcal{C}| \times |\mathcal{S}|)\), will approach close to 0 if $|\mathcal{C}|$ and $|\mathcal{S}|$ become very large. This implies that in practice, it is infeasible to utilize caching techniques for the \textit{bi-encoder} architecture if the sentences and conditions of interest are sufficiently diverse.}

\begin{table}[t!]
\begin{center}
\small
\setlength{\tabcolsep}{0.2em}
\begin{tabular}{lccc}
\toprule[1.5pt]
\textbf{Method} & \textbf{Time} & \textbf{HitRate} & \textbf{Cache} \\
\midrule
\multicolumn{4}{l}{\textit{bi-encoder architectures}} \\
\midrule
SimCSE\(_{base}\) & 791.71s & 1.46\% & 110.87MB  \\
SimCSE\(_{large}\) & 1498.65s & 1.46\% & 147.26MB \\
\midrule
\multicolumn{4}{l}{\textit{tri-encoder architectures}} \\
\midrule
SimCSE\(_{base}\) & 441.17s & 64.11\% & 60.57MB \\
SimCSE\(_{base+hyper64\text{-}cl}\) & 525.62s & 64.11\% & 2.17GB \\
SimCSE\(_{base+hyper\text{-}cl}\) & 541.55s & 64.11\% & 12.81GB \\
\addlinespace
SimCSE\(_{large}\) & 832.19s & 64.11\% & 80.45MB \\
SimCSE\(_{large+hyper64\text{-}cl}\) & 990.94s & 64.11\% & 3.82GB \\
SimCSE\(_{large+hyper\text{-}cl}\) & 960.84s & 64.11\% & 22.75GB \\
\bottomrule[1.5pt]
\end{tabular}
\end{center}
\caption{Analysis of inference time, cache hit rate, and memory usage for different architectures and methods on the entire C-STS dataset.}
\label{tab:tab3}
\end{table}

\begin{table}[t!]
\begin{center}
\small
\setlength{\tabcolsep}{0.2em}
\begin{tabular}{lccc}
\toprule[1.5pt]
\textbf{Method} & \textbf{Time} & \textbf{HitRate} & \textbf{Cache} \\
\midrule
\multicolumn{4}{l}{\textit{bi-encoder architectures}} \\
\midrule
SimKGC\(_{base}\) & 994.41s & 46.65\% & 295.29MB \\
SimKGC\(_{large}\) & 1806.18s & 46.65\% & 392.2MB \\
\midrule
\multicolumn{4}{l}{\textit{tri-encoder architectures}} \\
\midrule
SimKGC\(_{base+hadamard}\)  & 435.571s & 85.32\% & 121.86MB  \\
SimKGC\(_{base+concatenation}\)  & 449.46s & 85.32\% & 121.86MB  \\
SimKGC\(_{base+hyper\text{-}cl}\) & 448.955s & 85.32\% & 146.57MB \\
\addlinespace
SimKGC\(_{large+hadamard}\) & 781.45s & 85.32\% & 161.85MB \\
SimKGC\(_{large+concatenation}\) & 783.228s & 85.32\% & 161.85MB \\
SimKGC\(_{large+hyper\text{-}cl}\) & 774.41s & 85.32\% & 205.81MB \\
\bottomrule[1.5pt]
\end{tabular}
\end{center}
\caption{Analysis of inference time, cache hit rate, and memory usage for different architectures and methods on the entire WN18RR dataset.}
\label{tab:tab4}
\end{table}




\subsection{Analysis of Embedding Clusters}\label{clustering with conditioning networks}

In this subsection, we explore the impact of Hyper-CL on generating conditioned sentence representations by visualizing the computed embeddings and analyzing them using clustering tools.

\begin{figure}[t!]
    \centering
    \begin{adjustbox}{minipage=\columnwidth,scale=0.8}
    \begin{subfigure}{\columnwidth}
		\includegraphics[width=\columnwidth]{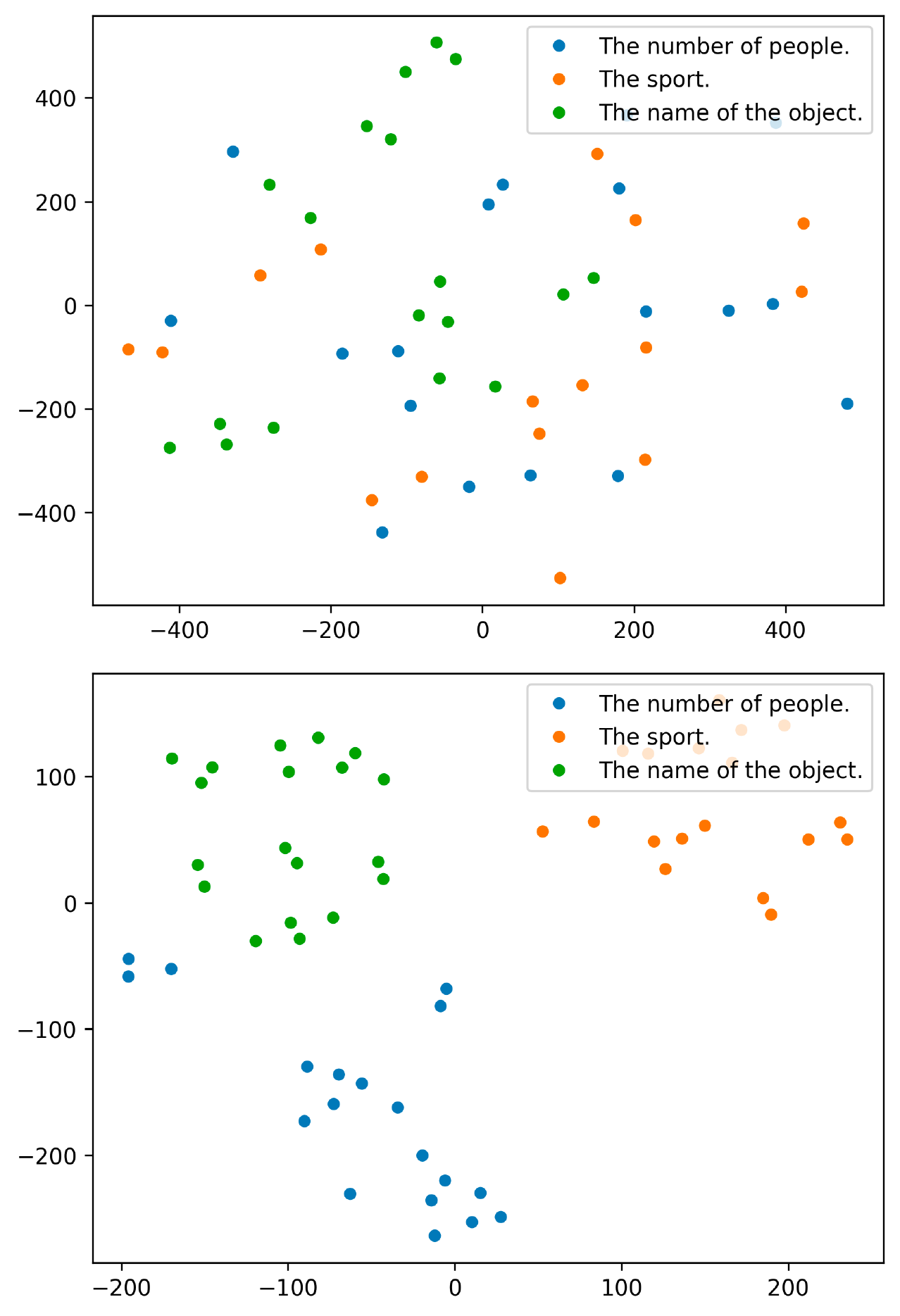}  
    \end{subfigure}\vspace{3mm}
    \end{adjustbox}
    \caption{Visualization of the clusters of sentence embeddings before (top) and after (bottom) projection onto condition subspaces by Hyper-CL.}
    \label{fig:fig_clustering}
\end{figure}

We visualize the vector space of sentence embeddings before and after transformation by $W_c$, which is the weight of a linear layer from the Hyper-CL's hypernetwork. 
This helps us observe if embeddings based on the same condition cluster after transformation, indicating proper differentiation based on conditions.
For this analysis, we choose three sets of 20 random sentences from the C-STS validation dataset, with sentences within each group sharing the same conditions.
The three conditions we select are: `The number of people', `The sport', and `The name of the object'.
As expected, Figure \ref{fig:fig_clustering} shows that sentence embeddings transformed with the same $W_{c_i}$ form clusters, meaning each embedding has projected to respective subspaces.

We complement the visual analysis with a quantitative evaluation. 
We perform K-means clustering on the sentence embeddings before and after transformation.\footnote{$K$ is set to 3, same as the number of chosen conditions.} 
Following the clustering, we compute the average impurity (entropy) of each sentence group, where a lower value suggests better conditioning of sentence embeddings.
Formally, the impurity $I$ based on the entropy of each (conditional) sentence  group $E(i)$ is given by:
\begin{equation*}
\resizebox{.99\linewidth}{!}{
    $I = \sum_i \frac{|C_i|}{|S|} E(i) = -\sum_i \frac{|C_i|}{|S|} \sum_j \frac{|L_{ij}|}{|C_i|} \log \frac{|L_{ij}|}{|C_i|}$},
\end{equation*}
where $|S|$ is the total number of sentences, $|C_i|$ is the number of sentences that should be labeled as condition $i$, and $|L_{ij}|$ represents the number of examples clustered as $j$ by K-means clustering within the $i^{th}$ (conditional) sentence group.
We discover that after projection done by Hyper-CL, $I$ changes from 0.739 to 0.270, indicating that Hyper-CL effectively projects sentence embeddings into distinct subspaces based on different conditions.

\subsection{Generalization Capabilities of Hyper-CL} \label{generalization capabilities of hyper-cl}
We examine the generalization capabilities of Hyper-CL, focusing on its ability to generalize to unseen conditions during training.
This enables a fine-grained evaluation of the conditioning ability and confirms its feasibility in realistic settings.

For the targeted study, we define two separate subsets of the C-STS validation set.\footnote{Due to the absence of publicly available gold-standard labels for the C-STS test set, we use the validation set for evaluation purposes. Additionally, we partition the C-STS training data in a 9:1 ratio to establish a new validation set.
Note also that we concentrate on C-STS in this section, as WN18RR and FB15K-237 do not provide `unseen' conditions.}
The first category is referred to as the `unseen' dataset, which consists only of data instances with conditions not present during training.
The second is named the `seen' dataset, which comprises data instances with conditions already seen in the training phase.
Statistically, the number of data instances for the `unseen' portion is 731 (25.79\% of the overall dataset), and for the `seen' portion, it is 2,103.

Experimental results on the `overall', `unseen', and `seen' datasets are listed in Table \ref{tab:tab5}.
For the embedding model $f$, we employed SimCSE$_{large}$.
Compared to the original \textit{tri-encoder} (the first row), Hyper-CL shows a clear performance improvement of 16-22 points in both `seen' and `unseen' settings, with particularly superior performance in the `unseen' setting. 
These findings highlight the superior generalization capabilities of Hyper-CL, enabling it to excel at handling unseen data.


\begin{table}[t]
\centering
\small
\begin{center}
\setlength{\tabcolsep}{0.2em}
\begin{tabular}{lccc}
\toprule[1.5pt]
\textbf{Method (Metric: Spearman)} & \textbf{Overall} & \textbf{Unseen} & \textbf{Seen}  \\
\midrule
SimCSE\(_{large}\)     & 32.13  & 13.93  & 25.02 \\
SimCSE\(_{large+hyper\text{-}cl}\)       & \textbf{38.59}  & \textbf{36.25}  & \textbf{41.14} \\
\bottomrule[1.5pt]
\end{tabular}
\end{center}
\caption{Generalization capabilities of Hyper-CL on the C-STS validation set. We compare the \textit{tri-encoder} baseline and Hyper-CL in both `unseen' and `seen' settings, using Spearman's correlation as the evaluation metric.}
\label{tab:tab5}
\end{table}

\subsection{Ablation Study on Contrastive Learning} \label{ablation}

In \cref{contrastive Learning in condition subspaces}, we argued that the joint utilization of hypernetworks and contrastive learning yields the best performance among the available options. 
To verify this, we evaluate four different variations of the \textit{tri-encoder} architecture on C-STS, whose details are as follows: (1) SimCSE$_{base}$: the \textit{tri-encoder} architecture trained only with $L_{MSE}$; (2) SimCSE$_{base+cl}$: the \textit{tri-encoder} trained with both $L_{MSE}$ and $L_{CL}$ but hypernetworks excluded; (3) SimCSE$_{base+hyper64}$: a variant of Hyper-CL ($K$=64) but trained only with $L_{MSE}$; (4) SimCSE$_{base+hyper64-cl}$: a normal Hyper-CL with low-rank approximation ($K$=64).
For a fair comparison, we ensure that the total number of parameters for each variant remains consistent, guaranteeing equal expressive power.

Table \ref{tab:tab6} shows that the contrastive learning objective ($L_{CL}$) is more effective when combined with hypernetworks.
This trend is clearly observed when comparing the performance increase from SimCSE\(_{base+hyper64}\) to SimCSE\(_{base+hyper64\text{-}cl}\) and that from SimCSE\(_{base}\) to SimCSE\(_{base+cl}\).

\begin{table}[t]
\small
\centering
\begin{center}
\begin{tabular}{lc}
\toprule[1.5pt]
\textbf{Method} & \textbf{Spearman} \\
\midrule
SimCSE\(_{base+hyper64\text{-}cl}\) & \textbf{37.96}  \\
SimCSE\(_{base+hyper64}\) & 35.38  \\
SimCSE\(_{base+cl}\) & 36.13  \\
SimCSE\(_{base}\) & 35.47  \\
\bottomrule[1.5pt]
\end{tabular}
\end{center}
\caption{Ablation study on the effectiveness of contrastive learning in condition subspaces. The results are from the C-STS validation set.}
\label{tab:tab6}
\end{table}

\subsubsection{Why is Contrastive Learning More Effective with Hypernetworks?}

The weight matrix $W_c = q(h_c)$, generated by the hypernetworks of Hyper-CL, is responsible for a linear transformation of a sentence embedding.
On the other hand, the Hadamard product of a sentence embedding and a condition embedding, which is computed in the original \textit{tri-encoder} architecture, can also be considered as a linear transformation, formulating the condition embedding $h_c$ as a diagonal matrix $W_{c'} = diag(h_c)$. 

To gauge the expressiveness of the two different transformations induced by $W_c$ and $W_{c'}$,
we calculate the variance of the Frobenius norm of these matrices during inference on a subset of the C-STS validation set.
For matrices with varying valid element counts (i.e., $W_c$ and $W_{c'}$), we normalize their Frobenius norm by dividing by the square root of the number of valid elements.
Experimental results show that the variance of the Frobenius norm, a measure of the matrices' expressive power, is significantly higher (0.0248) for Hyper-CL's transformations ($W_c$) compared to the Hadamard product (0.001; $W_{c'}$). 
These findings imply that hypernetworks endow the transformation with enhanced expressive power. Consequently, it is reasonable to expect that the contrastive learning process that leverages hypernetworks would also exhibit greater effectiveness.

\section{Conclusion}

We propose Hyper-CL, a method that combines hypernetworks with contrastive learning to generate conditioned sentence representations. 
In two representative tasks requiring conditioning on specific perspectives, our approach successfully narrows the performance gap with the \textit{bi-encoder} architecture while maintaining the time efficiency characteristic of the tri-encoder approach. 
We further validate the inner workings of Hyper-CL by presenting intuitive analyses, such as visualizations of the embeddings projected by Hyper-CL. 
In future work, we plan to explore a broader range of applications for Hyper-CL and to investigate its refinement.

\section*{Limitations} 
We have only explored applying our approach to encoder models, leaving room for applications on decoder models. Additionally, despite the variety of existing contrastive learning methodologies, we adhere to utilizing the contrastive learning objectives provided by the tasks.

\section*{Ethics Statement}
In this study, we utilized models and datasets publicly available from Huggingface. 
All datasets for evaluation are open-source and comply with data usage policies. 
However, some datasets (e.g., FB15k-237) are derived from Freebase, a large collaborative online collection that may contain inherently unethical information. 
We conducted a thorough inspection to check if our dataset contained any unethical content. 
No harmful information or offensive topics were identified during the human inspection process.

\section*{Acknowledgements}
This work was supported by Institute of Information \& communications Technology Planning \& Evaluation (IITP) grant funded by the Korea government(MSIT) (No.RS-2020-II201373, Artificial Intelligence Graduate School Program(Hanyang University)).
This work was supported by Institute of Information \& communications Technology Planning \& Evaluation (IITP) under the artificial intelligence semiconductor support program to nurture the best talents (IITP-2024-RS-2023-00253914) grant funded by the Korea government(MSIT).
This work was supported by the National Research Foundation of Korea(NRF) grant funded by the Korea government(*MSIT) (No.2018R1A5A7059549). *Ministry of Science and ICT.

\bibliography{custom}

\clearpage

\appendix

\section{Training Details}
\label{appendix:Training details}

In this section, we describe the hyperparameters used for training Hyper-CL on the two evaluation tasks employed in this paper. 
We implemented Hyper-CL using the Transformers package \cite{wolf-etal-2020-transformers}. 
For both tasks, Hyper-CL utilized the [CLS] token embedding computed by an encoder model as a sentence representation.

\textbf{Conditional Semantic Textual Similarity (C-STS)}: We conducted a hyperparameter search over learning rates $\in\{1e\text{-}5, 2e\text{-}5, 3e\text{-}5\}$, weight decays $\in\{0.0, 0.1\}$, and temperatures $\in\{1.0, 1.5, 1.7, 1.9\}$. The hyperparameter set yielding the best scores on the C-STS validation set with three random seeds was used for the final evaluation of the test set. 
As a result, we adopted the hyperparameters as shown in Table \ref{tab:tab7}.

\begin{table}[h]
\begin{center}
\small
\setlength{\tabcolsep}{0.5em}
\begin{tabular}{llcc}
\toprule[1.5pt]
\textbf{Method} & \textbf{LR} & \textbf{WD} & \textbf{Temp} \\
\midrule
DiffCSE$_{base+hyper\text{-}cl}$ & 3e-5 & 0.1 & 1.5 \\
DiffCSE$_{base+hyper64\text{-}cl}$ & 1e-5 & 0.0 & 1.5 \\
SimCSE$_{base+hyper\text{-}cl}$ & 3e-5 & 0.1 & 1.9 \\
SimCSE$_{base+hyper64\text{-}cl}$ & 2e-5 & 0.1 & 1.7 \\
SimCSE$_{large+hyper\text{-}cl}$ & 2e-5 & 0.1 & 1.5 \\
SimCSE$_{large+hyper85\text{-}cl}$ & 1e-5 & 0.1 & 1.9 \\
\bottomrule[1.5pt]
\end{tabular}
\end{center}
\caption{Hyperparameters determined for the C-STS task. 
The abbreviations \textbf{LR}, \textbf{WD}, \textbf{Temp} stands for learning rate, weight decay, and temperature, respectively.}
\label{tab:tab7}
\end{table}

\textbf{Knowledge Graph Completion (KGC)}: We utilized the same set of hyperparameters proposed in SimKGC \cite{wang-etal-2022-simkgc}.

\begin{table}[!b]
\begin{center}
\small
\begin{tabular}{lcc}
\toprule[1.5pt]
\multicolumn{1}{l}{\textbf{Method}} & \textbf{Rank} ($K$)        & \textbf{Spearman} \\
\midrule
\multirow{6}{*}{DiffCSE\(_{base+hyperK\text{-}cl}\)}        & 768  (=768/1)   & 33.82      \\
                                     & 192  (=768/4)    & 34.73  \\
                                     & 96  (=768/8)   & 34.16  \\
                                     & \textbf{64  (=768/12)}   & \textbf{33.10}      \\
                                     & 48  (=768/16)   & 33.31      \\
                                     & 32  (=768/24)   & 31.68  \\
\midrule
\multirow{6}{*}{SimCSE\(_{base+hyperK\text{-}cl}\)}         & 768  (=768/1)   & 38.75      \\
                                     & 192  (=768/4)    & 38.66  \\
                                     & 96  (=768/8)   & 35.69  \\
                                     & \textbf{64  (=768/12)}   & \textbf{38.36}      \\
                                     & 48  (=768/16)   & 37.02      \\
                                     & 32  (=768/24)   & 36.92  \\
\midrule
\multirow{6}{*}{SimCSE\(_{large+hyperK\text{-}cl}\)}       & 1024 (1024/1)   & 39.60       \\
                                     & 256 (=1024/4)    & 38.76   \\
                                     & 128 (=1024/8)    & 38.19  \\
                                     & \textbf{85 (=1024/12)}    & \textbf{38.12}  \\
                                     & 64 (=1024/16)    & 37.83                     \\
                                     & 42 (=1024/24)    & 37.44                     \\
                                     
\bottomrule[1.5pt]
\end{tabular}
\end{center}
\caption{Ablation study of different ranks ($K$).}
\label{tab:tab8}
\end{table}

\section{Ablation Study on the Selection of $\textbf{K}$}
\label{appendix:selection of k}

The selection of $K$ for constructing lightweight hypernetworks is closely related to the size of sentence embeddings. 
We empirically evaluated the validation set to determine suitable values for $K$ by dividing the embedding sizes of the base (768) and large (1024) encoder embeddings with various divisors (1, 4, 8, 12, 16, 24).
According to Table \ref{tab:tab8}, we observed that for SimCSE$_{large}$, the performance difference between K=128 and K=85 is just 0.07 points, while trainable parameters increase 1.5x.
In conclusion, we found that setting the divisor to 12 (resulting in K values of 64 and 85 for the base and large models, respectively) achieves an optimal balance between performance and efficiency.

\end{document}